\pdfoutput=1

\documentclass[11pt]{article}

\usepackage[]{EMNLP2022}

\usepackage{times}
\usepackage{latexsym}
\usepackage{amssymb,amsfonts}
\usepackage{amsmath}
\usepackage{algorithmic}
\usepackage{graphicx}
\usepackage{textcomp}
\usepackage{kotex}
\usepackage{booktabs}
\usepackage{multirow}
\usepackage{arydshln}
\usepackage{tabularx}
\usepackage{xcolor}
\usepackage{color}
\usepackage{amssymb}
\usepackage{mathtools}
\usepackage{makecell}
\usepackage{hyperref}
\usepackage{multicol}
\usepackage[subrefformat=parens]{subcaption}
\usepackage{lipsum, babel}
\usepackage{appendix}
\usepackage[symbol]{footmisc}

\usepackage[T1]{fontenc}

\usepackage[utf8]{inputenc}

\usepackage{microtype}

\usepackage{inconsolata}

\title{You Truly Understand What I Need \\ : Intellectual and Friendly Dialogue Agents grounding \\ Knowledge and Persona}

\author{Jungwoo Lim$^{1}$, Myunghoon Kang$^{1}$\footnotemark[1], Yuna Hur$^{1}$\footnotemark[1], Seungwon Jung$^{1}$\footnotemark[1], Jinsung Kim$^{1}$\footnotemark[1], \\ \textbf{Yoonna Jang$^{1}$, Dongyub Lee$^{3}$, Hyesung Ji$^{2}$, Donghoon Shin$^{2}$}, \\ \textbf{Seungryong Kim$^{1}$\footnote[4]{} \space and  Heuiseok Lim$^{1}$\footnote[4]{}} \\$^{1}$Korea University, $^{2}$Dialogue Tech Division, NCSOFT, $^{3}$Naver Corporation\\
\texttt{\small \{wjddn803,chaos8527,yj72722,redlion0929,jin62304,seungryong\_kim,limhseok\}@korea.ac.kr}, \\
\texttt{\small \{hyesung84,dhshin\}@ncsoft.com}, \texttt{\small dongyub.lee@navercorp.com}
}

\begin{document}
\maketitle
\begin{abstract}
To build a conversational agent that interacts fluently with humans, previous studies blend knowledge or personal profile into the pre-trained language model. However, the model that considers knowledge and persona at the same time is still limited, leading to hallucination and a passive way of using personas. We propose an effective dialogue agent that grounds external knowledge and persona simultaneously. The agent selects the proper knowledge and persona to use for generating the answers with our candidate scoring implemented with a poly-encoder. Then, our model generates the utterance with lesser hallucination and more engagingness utilizing retrieval augmented generation with knowledge-persona enhanced query. We conduct experiments on the persona-knowledge chat and achieve state-of-the-art performance in grounding and generation tasks on the automatic metrics. Moreover, we validate the answers from the models regarding hallucination and engagingness through human evaluation and qualitative results. We show our retriever's effectiveness in extracting relevant documents compared to the other previous retrievers, along with the comparison of multiple candidate scoring methods. Code is available at \url{https://github.com/dlawjddn803/INFO}

\end{abstract}
\footnotetext{\footnote[1]{} ~~Equal Contributors}
\footnotetext{\footnote[4]{} Corresponding author}
\section{Introduction}

\begin{table}[t]
\centering
\resizebox{0.45\textwidth}{!}{%

\begin{tabular}{@{}llll@{}}
\toprule
\multicolumn{4}{l}{\textbf{Dialogue}} \\
\midrule
\multicolumn{4}{l}{
\begin{tabular}[c]{p{10cm}}
\textit{Human}:~Is it in England?\\ 
\textit{Machine}:~No, it is actually in Scotland where you are going.  \\
\textit{Human}:~Where in Scotland?
\end{tabular}} \\ \midrule
\multicolumn{2}{l}{\textbf{Human's Persona}}\\ \midrule
\multicolumn{2}{l}{
\begin{tabular}[c]{p{9.5cm}}
\textcolor{red}{I will travel through North Ayrshire.} \\
I am going to Scotland.\\
I like history.\\
I am interested in architecture.\\
I love to garden.\\ 
\end{tabular}} \\ \midrule
\multicolumn{4}{l}{\textbf{Ground Truth Knowledge}} \\ \midrule
\multicolumn{4}{l}{\begin{tabular}[c]{p{10cm}}\textcolor{blue}{Eglinton Castle was a large Gothic castellated mansion in Kilwinning, North Ayrshire, Scotland.}\end{tabular}} \\ \midrule
\multicolumn{4}{l}{\textbf{Predicted Answers}}\\ 

\midrule
~~~BART$_{base}$  & \multicolumn{3}{l}{\begin{tabular}[c]{p{7.2cm}}It is in Scotland, which is a place you love.\end{tabular}}\\ 
\cdashline{1-4}[0.8pt/1.2pt]
~~~BART$_{large}$ & \multicolumn{3}{l}{\begin{tabular}[c]{p{7.2cm}}It is in Scotland. in Scotland. in Scotland. in\end{tabular}}\\ 
\midrule
\multicolumn{4}{l}{\textbf{Ground Truth Response}}\\ \midrule
\multicolumn{4}{l}{\begin{tabular}[c]{p{10cm}}
\textcolor{blue}{It is in North Ayrshire} so \textcolor{red}{you could visit when you travel through.}
\end{tabular}} \\ 
\bottomrule
\end{tabular}%
}
\caption{Example of the generated answers from a typical generative model, i.e., BART. We can find that BART$_{base}$ uses different persona sentence which has not appeared human's personal profiles resulting in hallucinated answer. Also, BART$_{large}$ generates less engaging answers by making use of the knowledge only to answer the question. Both generated responses are in the situation of hallucination and are less engaging. }
\label{tab:example}
\end{table}


To build an ultimate conversational agent that interacts with humans fluently, previous studies provide generative neural network-based models~\citep{sordoni2015neural,vinyals2015neural}. Although the answers generated from those models are plausible, they lack informativeness and engagingness resulting in bland responses compared to humans~\citep{li2016diversity,gao2018neural}. However, for knowledgeable and attractive conversation, people usually provide informative replies by considering the background of the person whom they are talking to. Towards a human-like manner of dialogue, \citet{ghazvininejad2018knowledge} and \citet{dinan2018wizard} introduce the knowledge-grounded conversation for the knowledgeable and informative responses, whereas \citet{zhang2018personalizing} suggest the persona-grounded dialogue for the personalized responses to the users.

To improve the machine's answer with the external knowledge base, one injects the factual knowledge into the parameters of the language model~\citep{raffel2020exploring,roberts2020much}. Despite the models' capability of utilizing external knowledge implicitly, they produce ``hallucinations'' in the responses~\citep{marcus2020next}. The hallucination in the dialogue involves the situation where the generated output contradicts the reference knowledge. Also, it includes the situation when the generated output cannot be confirmed from the knowledge source~\citep{ji2022survey}. To mitigate these hallucinated answers, hybrid models employing parametric memory with non-parametric (i.e., retrieval-based) memory are introduced to directly access external memories, leading the source to be inspected and interpreted~\citep{karpukhin2020dense,petroni2020context,lewis2020retrieval}. 

On the other hand, \citet{zhang2018personalizing} suggest persona-chat dialogues with the corresponding personal profiles of each interlocutor to avoid general and monotonous answers from the machine. Though \citet{see2019makes,liu-etal-2020-personachat} show comparable quality in generating personalized conversation, the generated utterances merely confirm each interlocutor's persona resulting in a passive manner of speaking such as ``I have four children''. In addition, the incoherent topics of the dialogues lead to shallow levels of conversation between the interlocutors. To elaborate on this chit-chat conversation supported by external knowledge, \citet{jang2022call} presents a novel persona-knowledge chat with a generative model that considers persona information and world knowledge altogether. Despite obtaining the knowledge and persona when generating the answers, the generative models' responses still exhibit both hallucination and lesser engagingness as in Table~\ref{tab:example}.

In this paper, we propose INFO (Intellectual and Friendly dialOg agents) that responds with external knowledge and persona simultaneously. Owing to the enhanced capturing relevancy between the context and each candidate set, the knowledge selector and persona selector for the grounding task are implemented with the poly-encoder. To alleviate hallucinated responses from the model, we adopt retrieval-augmented generation (RAG)~\citep{lewis2020retrieval} by utilizing non-parametric memory and parametric generator in addition to the enhanced input query. By injecting predicted sources as input to the retrieved-augmented generator, our model maintains consistency between grounding and generation while training. Therefore, our model generates more knowledgeable and engaging answers in an active manner with less hallucination. 

We show that INFO achieves the highest scores on both grounding and generation tasks in empirical experiments. Also, we compare diverse candidate scoring modules including bi-encoder, cross-encoder, and poly-encoder and demonstrate their effect on generation. We additionally conduct experiments to show the effectiveness of the retriever module compared to sparse and dense retrievers. The qualitative results and human evaluation are also presented to validate our model's capability to generate human-like answers. 

Our contributions are as follows:
\begin{itemize}
\item {We propose the model that grounds persona information and external knowledge with lesser hallucination and adequate utilization of persona in an active manner simultaneously.}

\item {Our approach suggests that the generated responses from the model are interpretable regarding what the model refers to while generating.}

\item {We show that INFO achieves the SoTA performance in all of the automatic metrics and demonstrate its comparable quality with human evaluation and qualitative analysis.}

\end{itemize} 

\begin{figure*}[t]
	\centering
	\includegraphics[width=0.9\textwidth]{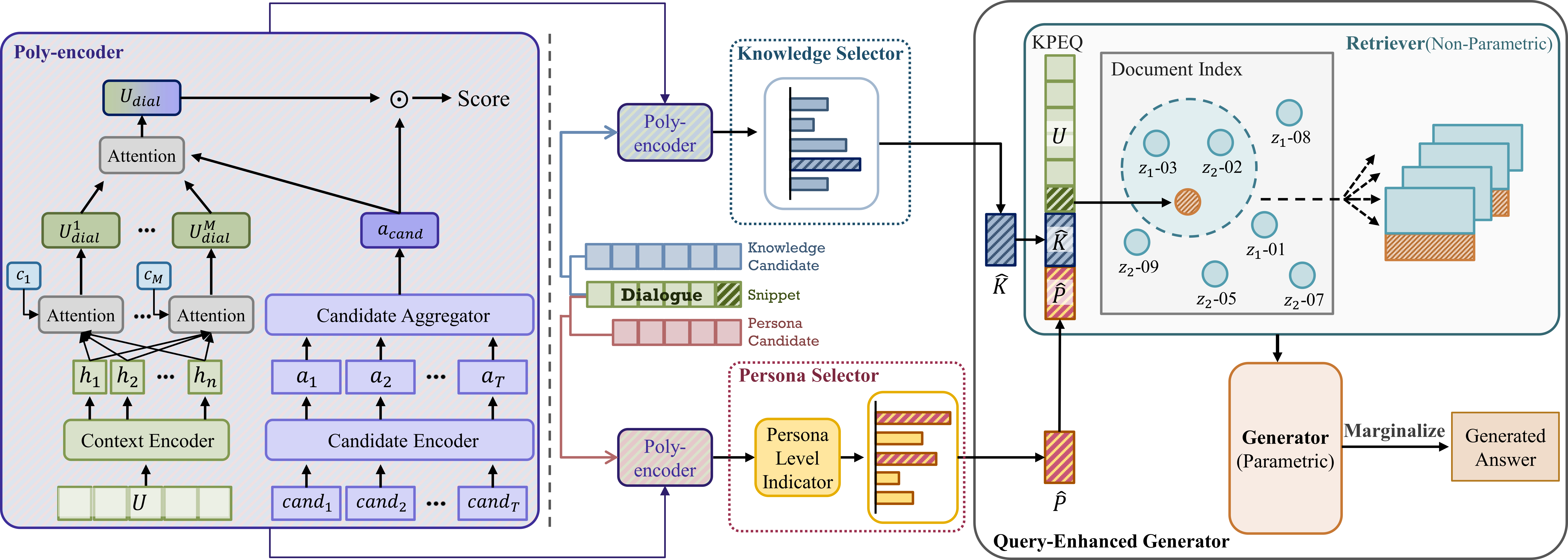}
\caption{Overview of our method. $U$ is the input comprises dialogue history and knowledge snippet, and $cand$ denotes each candidate from the grounding tasks. The grounding score is obtained through the dot product operation with the representation of input context $U_{dial}$ and candidate $a_{t}$. The predicted sources convert into the knowledge-persona enhanced query (KPEQ) with dialogue history and KPEQ is fed into the retrieval-augmented generator to generate the responses.} \label{fig:overview}
\end{figure*}

\section{Related Works}

\subsection{Knowledge Grounded Conversation}
To let the neural network models ground external knowledge and generate informative answers, \citet{ghazvininejad2018knowledge} suggests a data-driven neural conversational agent that provides knowledgeable answers. Also, \citet{dinan2018wizard} introduces open-domain dialogue where the two speakers are talking with Wikipedia knowledge. To inject the external knowledge into the pre-trained language model efficiently, \citet{raffel2020exploring,roberts2020much} success in equipping the knowledge into the parameters and show comparable performance in open-domain question and answering tasks. However, the approach is not capable of expand or revise their inherent knowledge and provides hallucination~\citep{marcus2020next}. To overcome the limitations, \citet{lewis2020retrieval} combines a pre-trained parametric model and non-parametric memory for the open-domain question and answering to reduce hallucination. Since their non-parametric memory can be updated without extra pre-training, revising knowledge is more efficient. Furthermore, it is found that a retrieval-augmented generator also reduces hallucination in knowledge-grounded conversation as well~\citep{shuster2021retrieval}, and a similar approach recently achieves outstanding performance in knowledge-grounded conversation~\citep{paranjape2021hindsight}.

\subsection{Persona Grounded Conversation}
In order to alleviate bland and general answers with consistent personality, \citet{zhang2018personalizing} constructs a persona-chat dataset. In the dataset, the two interlocutors chat with the persona profile sentences. Along with this dataset, \citet{zhang2018personalizing} introduces the model with a profile memory network by considering the dialogue history to perform attention over the persona. They enlarge the persona-chat dataset with Reddit corpus, and pre-trained the model with these dataset. After that, they fine-tune pre-trained model on the persona-chat~\citep{mazare2018training}. Also, \citet{liu-etal-2020-personachat} trains a receiver to reinforce the mutual persona understanding between interlocutors, and  \citet{wolf2019transfertransfo} utilize pre-trained models~\citep{radford2019language} to build personalized dialogue agents.

\subsection{Encoders for Sentence Scoring}

There exist diverse encoder structures for sentence scoring. Bi-encoder scores the relevance between sentences by feeding context and candidates into separate encoders. An example of bi-encoders are memory networks ~\citep{zhang2018personalizing}, transformer memory networks~\citep{dinan2018wizard}, LSTM~\citep{lowe2015ubuntu}. Since bi-encoder calculates with cached encoded sentence representations, it is relatively fast in computation. However, the bi-encoder has a limitation of capturing mutual information between context and candidates. Cross-encoder, on the other hand, scores by aligning context and candidates in one sequence. A type of cross-encoders is a sequential matching network that is based on deep matching networks~\citep{yang2018response} and gated self-attention~\citep{zhang2018modeling}. Although using a cross-encoder can achieve rich interaction between the sentences within the encoder, the problem of slow processing still remains. To exploit both benefits of each model, poly-encoder adopts attention mechanism into the bi-encoder architecture and shows satisfactory performances as cross-encoder with fast inference time~\citep{humeau2019poly}. For the enhanced representation of grounding knowledge and persona, we employ a poly-encoder as a selector for each grounding task. 

\section{Method}
\label{sec:method}
To generate more knowledgeable and engaging dialogue, we introduce our conversational model that grounds external knowledge and persona information as in Figure \ref{fig:overview}. We first encode the input with the pre-trained language model, and then choose the proper knowledge and persona from the given candidates for each selector. We employ poly-encoder
~\citep{humeau2019poly} as knowledge selector and persona selector to exploit its enhanced capability of capturing relevance between candidate set and context (i.e., dialogue history). Then, the predicted persona and knowledge are aligned into one sequence to the dialogue history for consistency between grounding and generation. The sequence is defined as a knowledge-persona enhanced query (KPEQ), then it feeds into the retriever-augmented generator (RAG). The generator then extracts the relevant paragraphs to refer from the knowledge index to reduce hallucination.

\subsection{Input Construction}
The given dialogue is notated as $\{(u_{1}^{hm}, u_{1}^{mc}), ... (u_{o}^{hm}, u_{o}^{mc})\}$, where $o$ is the number of rounds. $u^{hm}$ and $u^{mc}$ indicate the utterances of human and machines, respectively. We first take $o$-th round dialogue history, except for the final machine's reply $u_{o}^{mc}$, for the initial input for the model. We define the clue of the dialogue as knowledge snippet $cl_{k}$ to inform the machine of which topic the user is interested in. The knowledge snippet is the name of the landmark that the user encounters, which is given topic from the dialogue. We then align the dialogue history and knowledge snippet into the one sequence for the model input as $ U = \{u_{1}^{hm}, u_{1}^{mc}, ... u_{o}^{hm}, cl_{k}\}$.

\subsection{Model Components}

\subsubsection{Poly-Encoder Based Candidate Scoring}
For knowledge and persona grounding tasks, we suggest poly-encoder-based candidate scoring to leverage the capability of capturing the semantic similarities between the context input and the candidates. It is employed to select proper sources to be used when generating the utterance. When the context input $U$ comes in, we compute the grounding scores of each candidate utilizing the embeddings of context input and encoded candidates in the poly-encoder. The grounding score is used to select the most suitable source(s) in the knowledge selector and persona selector, which will be introduced in the following Section \ref{sec:ks} and \ref{sec:ps}.

In poly-encoder architecture~\citep{humeau2019poly}, candidates are fed into the candidate encoder and denoted as $\{a_{1}, ..., a_{T}\}$ where $T$ is the number of candidates in the set. Each candidate embedding $a_{t}$ is the first output of the candidate encoder, which is represented by the transformer model. After encoding candidates, the context input (i.e., dialogue history) is embedded with a separate context encoder. Unlike the candidate encoder, the context encoder embeds the dialogue into multiple vectors through $M$ context codes $\{{c_{1}, ... c_{M}\}}$, which are learned for capturing diverse aspects of a given context rather than using one embedding. Each context code is used to extract $U_{dial}^{m}$ by attending over all the previous layer's output as follows. 

\begin{equation}
{
     U_{dial}^{m} = \sum_{j}^{} w_{j}^{c_{m}}h_{j} 
     \label{equation:poly1}
} 
\end{equation} 
Note that the $h_{1}, ..., h_{n}$ is the output of the pre-trained language model and $n$ is the number of tokens in the input. The weights are computed as  $(w_{1}^{c_{m}}, ..., w_{n}^{c_{m}}) =$  $\mathbf{\mathrm{softmax}} (c_{m}\cdot h_{1}, ..., c_{m} \cdot h_{n})$.

Then, the final attention proceeds between the global features of the input and a given candidate. In other words, the final dialogue feature $U_{dial}$ is obtained by aggregating each dialogue feature $U^{m}_{dial}$, while gaining richer interactions with context codes as in Equation \ref{equation:poly2}.
\begin{equation}
{
     U_{dial} = \sum_{m}^{} w_{m} U_{dial}^{m},
     \label{equation:poly2}
} 
\end{equation} 
where $w_{1}, ..., w_{M}$ can be obtained from $\mathbf{\mathrm{softmax}} (a_{t} \cdot U_{dial}^{1}, ..., a_{t} \cdot U_{dial}^{M})$.

The final predicted candidate is chosen based on the highest score that is acquired from the dot product operation as ($U_{dial} \cdot a_{t}$).

\subsubsection{Knowledge Selector (KS)} \label{sec:ks}
We build a knowledge selector for the knowledge grounding task, employing poly-encoder-based candidate scoring. When the grounding scores are produced from the candidate scoring module, the label with the highest score is selected as the predicted knowledge. 

The knowledge loss $\mathcal{L}_{KG}$ for the knowledge grounding task is computed with cross-entropy loss~\citep{brier1950verification} as in Equation \ref{equation:kloss}.
\begin{equation}
{
     \mathcal{L}_{KG} = -\sum_{j}^{} kl_{j} \cdot \mathrm{log} \hat{kl_{j}}
     \label{equation:kloss},
} 
\end{equation} 
$kl_{j}$ is the ground-truth label from the knowledge candidates of the $j$-th example.

\subsubsection{Persona Selector (PS)}\label{sec:ps}
We also implement a persona selector for the persona grounding task. Since multiple personas can be chosen to generate the responses, consideration of one or more persona sentences are needed. Similar to the knowledge selector, we assign the grounding score to each persona candidate with the candidate scoring module as in Equation \ref{equation:poly1} and \ref{equation:poly2}. 

When the scores of each candidate are computed from the candidate scoring module, then the persona level indicator classifies which the number of the persona should be selected with the \texttt{[CLS]} token of the model input $U$. After predicting the level of persona-engagingness, we pick persona sentences to be grounded according to the number predicted. For example, if the persona level indicator predicts 2, then top-2 persona sentences are chosen in the persona grounding task. The selected persona sentence(s) are marked as 1 otherwise, 0. We use binary cross-entropy loss for persona grounding as in Equation \ref{equation:ploss}.

\begin{equation} \label{equation:ploss}
{   
    \begin{aligned}
    \mathcal{L}_{PG} & = \\ & -\sum_{j}^{} pl_{j} \cdot \mathrm{log} \hat{pl_{j}} + (1-pl_{j}) \cdot \mathrm{log} (1-\hat{pl_{j}})
    \end{aligned}
} 
\end{equation} 

Note that $pl_{j}$ is the ground-truth label from the knowledge candidates of the $j$-th example.

\subsubsection{Query-Enhanced Generator}
Following the works of \citet{lewis2020retrieval}, we exploit the retrieval augmented generation's capability to reduce hallucination and access the memory directly. For a consistent way of training while solving grounding and generation tasks, we reconstruct the query that feeds into the retriever. When the knowledge and persona are predicted from each selector, we aggregate them with dialogue history into one sequence. Then, the final query is denoted as $\mathrm{KPEQ} = \{U; \hat{P}; \hat{K}\}$ and defined as a knowledge-persona enhanced query. $\hat{P}$ and $\hat{K}$ are predicted persona and knowledge from each candidate set, respectively. 

The retriever $r_{\eta}$ aims to search top-K latent paragraphs with the KPEQ. We utilize a pre-trained dense passage retriever (DPR)~\citep{karpukhin2020dense} trained on natural question dataset~\citep{kwiatkowski2019natural} which has parametric memory and bi-encoder architecture to retrieve a latent document embedding following \citet{lewis2020retrieval} :
\begin{equation} \label{equation:retriever1}
{
     r_{\eta}(z|\mathrm{KPEQ}) \varpropto exp (\mathbf{d}(z) ^{\top} \mathbf{q}(\mathrm{KPEQ})),
} 
\end{equation} 

where $\mathbf{d}(\cdot)$ is an embedding from a document encoder and $\mathbf{q}(\cdot)$ is a representation from query encoder, both implemented with BERT$_{base}$. $z$ denotes the list of document.

With the relevant paragraphs from the retriever, we employ RAG-Token architecture as the generator to borrow its strength of predicting each target token based on top-K different paragraphs. Since RAG-Sequence, which has a different architecture to RAG-Token, uses the same document from the retriever to predict each token as depicted in Equation \ref{equation:generator-seq}, the result may opt to depend on the retrieved document~\citep{lewis2020bart}. The two different versions of RAGs~\citep{lewis2020retrieval} are as follows: 

\begin{equation} \label{equation:generator-seq}
{
    \begin{aligned}
    S_{\text{RS}}(y|x) & \approx \\ & \sum_{\mathclap{z \in \text{top-k}(p(\cdot|x))}}  r_{\eta}(z|x) \prod_{i}^{N} g_{\theta}(y_{i}|x, z, y_{1:i-1})
    \end{aligned}
} 
\end{equation}
\begin{equation} \label{equation:generator-tok}
{
    \begin{aligned}
    S_{\text{RT}}(y|x) & \approx \\ & \prod_{i}^{N}~~~~~ \sum_{\mathclap{z \in \text{top-k}(p(\cdot|x))}}  r_{\eta}(z|x) g_{\theta}(y_{i}|x, z, y_{1:i-1}),
    \end{aligned}
} 
\end{equation} 
where S$_{RS}$ indicates our method with RAG-Sequence architecture and S$_{RT}$ denotes ours with the RAG-Token model. $x$ is a token of $\mathrm{KPEQ}$ and $y_{i}$ is a single token from the ground truth responses. Also, $z$ is a retrieved paragraph from the retriever and $N$ is the maximum sequence length.

\begin{table*}[t]
\centering
\scalebox{0.85}{
\begin{tabular}{lccccccccc}
\toprule
\multicolumn{1}{c|}{\multirow{2}{*}{Models}} & \multicolumn{6}{c|}{Generation}                       & \multicolumn{2}{c}{Grounding (Acc.)} \\ \cline{2-9} 
\multicolumn{1}{c|}{}  
& \multicolumn{1}{c}{chrF++}  & \multicolumn{1}{c}{BLEU} & \multicolumn{1}{c}{R-1} & \multicolumn{1}{c}{R-2} & \multicolumn{1}{c}{R-L} & \multicolumn{1}{c|}{BERTScore} & \multicolumn{1}{c}{Persona}          & \multicolumn{1}{c}{Knowledge}         \\ \midrule
\midrule
\multicolumn{1}{l|}{GPT2$_{small}$} & 28.73 & 11.43 & 36.58 & 19.44 & 32.62 & \multicolumn{1}{c|}{88.56}  & 67.44 & 69.59 \\
\multicolumn{1}{l|}{GPT2$_{medium}$} & 30.12 & 12.31 & 38.29  & 21.17 & 34.12 & \multicolumn{1}{c|}{88.92} & 67.44 & 72.42 \\ \midrule
\multicolumn{1}{l|}{BART$_{base}$} & 29.77 & 11.99 & 36.24 & 19.73 & 32.13 & \multicolumn{1}{c|}{88.35} &  67.45 & 72.18 \\
\multicolumn{1}{l|}{BART$_{large}$} & 30.69 & 11.91 & 36.57 & 19.83 & 32.05 & \multicolumn{1}{c|}{88.10} & 67.44 & 71.01 \\ \midrule
\multicolumn{1}{l|}{\textbf{INFO} ($S_{RS}$)} & 51.33 & 29.36 & 53.36 & 40.36 & 51.16 &  \multicolumn{1}{c|}{92.00} & \textbf{82.70} & \multicolumn{1}{c}{\textbf{99.24}} \\
\multicolumn{1}{l|}{\textbf{INFO} ($S_{RT}$)} & \textbf{53.29} & \textbf{31.46} & \textbf{58.26} & \textbf{42.35} & \textbf{53.06} &  \multicolumn{1}{c|}{\textbf{92.29}} & 80.87 & \multicolumn{1}{c}{99.22} \\

\bottomrule
\end{tabular}
}
\caption{Main results on the official validation set. $S_{RS}$ denotes our method with RAG-Sequence architecture and $S_{RT}$ indicates the model with RAG-Token model as generator. The models are evaluated by generation metrics, including chrF++, BLEU, ROUGE-1 (R-1), ROUGE-2 (R-2), ROUGE-L (R-L), and BERTScore. \label{tab:main_results}}
\end{table*}

The $S_{RT}$ generator $g(\cdot)$ marginalizes the loss from different paragraphs when generating answers. In detail, the generator outputs a distribution for the next token for each document before marginalizing as in Equation \ref{equation:generator-tok} where $\eta$ denotes the parameter of the retriever, and $\theta$ indicates the parameter of the generator. After that, the generator repeats the process with the following output token. 
Finally, the $S_{RT}$ aims to generate the next token following an auto-regressive manner with a standard beam search. In other words, the model minimizes the negative marginal log-likelihood for each input/output pair $(\mathrm{KPEQ}_{j}, y_{j})$. The language model loss is formulated as :

\begin{equation} \label{equation:sloss}
\begin{aligned}
\mathcal{L}_{S} = - \sum_{j} \mathrm{log} p(y_{j}|\mathrm{KPEQ}_{j}) 
\end{aligned}
\end{equation}

\subsection{Final Objectives}
We then train the full model in the multi-tasking manner. The full objectives of the model is indicated as Equation \ref{equation:full_obj}. 
\begin{equation} \label{equation:full_obj} 
{
    \begin{aligned}
    \mathcal{L} = \lambda_{KG}\mathcal{L}_{KG} + \lambda_{PG}\mathcal{L}_{PG} + \lambda_{S}\mathcal{L}_{S}
    \end{aligned}
} 
\end{equation} 

We control the proportion of each task and we set $\lambda_{KG}$, $\lambda_{PG}$, and $\lambda_{S}$ as 1:1:5 for the experiments, respectively. We find the value of each $\lambda$ with manual search.

\section{Experiments}
\subsection{Experiment Details}
\paragraph{Dataset}
FoCus \cite{jang2022call} is the dataset for customized dialogue benchmark, where each conversation is directly grounded with knowledge and persona. The dataset includes knowledge-aware dialogue with personal profiles between humans and machines. There are 12,484 dialogues about 5,152 knowledge sources from Wikipedia and 32,855 persona sentences. To validate the knowledge grounding capability and customized dialogue generation, we evaluate our method with the official FoCus validation set for the effectiveness of experiments since the result from the official test set can be tested only through the leaderboard\footnote{https://codalab.lisn.upsaclay.fr/competitions/3754}.

\paragraph{Experimental Setup}
For each candidate scoring module, we implement  poly-encoder~\citep{humeau2019poly} with BERT$_{large}$, and the number of context codes is 16. For the dialogue generation, we implement our method with Hugging Face~\citep{wolf-etal-2020-transformers} and use \texttt{facebook/rag-token-nq} as the backbone model. We use the same architecture of retriever and generator from RAG along with the decoding and leverage our knowledge index for non-parametric query-document ranking with FAISS library~\citep{johnson2019billion}. The knowledge index consists of the paragraphs from the given Wikipedia knowledge entitled with the name of the given landmark. We set learning rate as 6.25e-6 with AdamW~\citep{kingma2014adam} for the optimization. The batch size is set as 32, and the number of dialogue history is 1. The whole model was trained for three epochs on RTX A6000 GPU and took 8 hours per one epoch.

\paragraph{Baselines}
We implement the baselines from previous study~\citep{jang2022call} and we conduct experiments with GPT-2~\citep{radford2019language} and BART~\citep{lewis2020bart} as well.  For a fair comparison, we demonstrate the results on GPT-2$_{small}$, which has 12 layers, and BART$_{base}$, which has 6 encoders and 6 decoder layers. Also, GPT-2$_{medium}$ contains 24 layers of the decoder, and BART$_{large}$ possesses 12 layers for each encoder and decoder.

\begin{table*}[t]
\centering
\scalebox{0.70}{
\begin{tabular}{ll|cccccc|ccc}
\toprule
\multicolumn{2}{c|}{\multirow{3}{*}{Model}}                       & \multicolumn{6}{c|}{Generation}           & \multicolumn{3}{c}{Grounding} \\ \cline{3-11} 
\multicolumn{2}{l|}{\multirow{2}{*}{}}                                                  & \multicolumn{1}{c}{\multirow{2}{*}{chrF++}} & \multicolumn{1}{c}{\multirow{2}{*}{BLEU}} & \multicolumn{1}{c}{\multirow{2}{*}{R-1}} & \multicolumn{1}{c}{\multirow{2}{*}{R-2}} & \multicolumn{1}{c}{\multirow{2}{*}{R-L}} & \multicolumn{1}{c|}{\multirow{2}{*}{BERTScore}} & \multicolumn{1}{c}{\multirow{2}{*}{\begin{tabular}[c]{@{}l@{}}Persona\\~(Acc.)\end{tabular}}} & \multicolumn{1}{c}{\multirow{2}{*}{\begin{tabular}[c]{@{}l@{}}Persona\\~~~(F1)\end{tabular}}} & \multicolumn{1}{c}{\multirow{2}{*}{\begin{tabular}[c]{@{}l@{}}Knowledge\\~~~(Acc.)\end{tabular}}}   \\
\multicolumn{2}{l|}{} & \multicolumn{6}{l|}{} & \multicolumn{3}{l}{}\\
\midrule
\midrule
\multicolumn{1}{l|}{\multirow{3}{*}{\begin{tabular}{p{0.5cm}}$S_{RT}$\end{tabular}}} & Bi-encoder         & 51.83 & 29.51  & 56.35 & 40.80 & 51.37 & 91.86 & \textbf{88.10} & 38.20 & 99.18                \\ \cdashline{2-11}[0.8pt/1.2pt]

\multicolumn{1}{l|}{} & Cross-encoder & 49.90 & 27.18 & 53.57 &  38.25  & 49.29 &  91.52  & 87.09 & 35.32 &  \textbf{99.49} \\ \cdashline{2-11}[0.8pt/1.2pt]
\multicolumn{1}{l|}{} & Poly-encoder & \textbf{53.29} & \textbf{31.46}  & \textbf{58.26} & \textbf{42.35} & \textbf{53.06} & \textbf{92.29} & 80.87 &  \textbf{39.56} & 99.22 \\
\bottomrule
\end{tabular}
} \caption{Performances comparison between the encoding modules for grounding tasks\label{tab:encoder_results}}
\end{table*}

\subsection{Automatic Evaluation}
We show the main results on the FoCus dataset with automatic metrics in grounding and generation tasks. The official metrics for the benchmark are chrF++~\citep{popovic-2017-chrf}, BLEU~\citep{Papineni02bleu:a}, ROUGE-1, ROUGE-2, and ROUGE-L~\citep{lin-2004-rouge}. To consider the semantic similarity score for each token between candidate and reference sentences using contextual representation, we additionally adopt BERTscore~\citep{bert-score}. For grounding task, we used accuracy for both knowledge and persona grounding, and F1 score for the persona grounding.

In Table \ref{tab:main_results}, it is found that our method shows substantial improvements in all the metrics from generation to grounding compared to the baselines. Especially, the performances of INFO increase over 18\% at least regarding the generation metrics except for BERTScore. Furthermore, our model achieves remarkable success in persona and knowledge accuracy. Unlike the performance in other generation metrics, $S_{RS}$ demonstrates better persona accuracy than $S_{RT}$. This result might be attributed to the architecture of the generator, which is more applicable to sentence classification tasks such as persona grounding. The official test result is also demonstrated in Appendix \ref{sec:appen_main}, but BERTscore is missing due to the unreleased ground truth.

\subsection{Human Evaluation}

We conduct a human evaluation to validate the responses from our model through Amazon Mturk services\footnote{https://www.mturk.com/}. The assessment criteria are fluency, adequacy, provenance, engagingness, and hallucination. In specific, provenance is the level of utilization of the ground truth knowledge into the responses, whereas engagingness means how much the answers are persona-related. Also, hallucination indicates whether the answer contradicts the persona and knowledge or cannot be verified from the source content. We randomly chose 50 dialogues from the official test set, and three workers were allocated to evaluate each dialogue generated by our model and baselines. We asked the workers to rank the answers according to each criterion following \citet{cho-may-2020-grounding}. Rank is scaled from 1 to 5, and the lower number is mapped to the better quality except for hallucination. The agreement between the annotators is calculated with Fleiss’ Kappa coefficient and is 0.4185 indicating fair agreement. The relations between the annotators hardly exist since we collect the results from the Amazon Mturk workers.

\begin{table}[h]
\centering
\scalebox{0.82}{
\begin{tabular}{l|ccccc}
\toprule
\multicolumn{1}{c|}{\multirow{2}{*}{Models}} & \multicolumn{5}{c}{Avg. Rank}  \\ 
\cline{2-6} & Ad. $\downarrow$ & Fl. $\downarrow$ & Prov. $\downarrow$ & Eng. $\downarrow$ & Hall. $\uparrow$ \\ \midrule
\midrule
GPT-2$_{small}$                  & 3.57     & 3.41    & 3.58        & 3.46       & 2.49 \\
GPT-2$_{medium}$                  & 3.11      &  3.10   & 3.04      & 3.25        & 3.02 \\ \cdashline{1-6}[0.8pt/1.2pt]
BART$_{base}$                   & 3.43      & 3.29     & 3.47        & 3.22       & 2.45 \\
BART$_{large}$                   & 3.31      & 3.63     & 3.29        & 3.44        & 2.69 \\ \cdashline{1-6}[0.8pt/1.2pt]
\textbf{INFO} (Ours)        & \textbf{1.57} & \textbf{1.57} & \textbf{1.62} & \textbf{1.63}   & \textbf{4.35} \\
\bottomrule
\end{tabular}
}
\caption{Human evaluation. The value in the table is the average rank of the each model's response. The abbreviation Ad.  Fl. Prov. Eng. and Hall denote adequacy, fluency, provenance, engaginess, and hallucination, respectively. \label{tab:human_eval}}
\end{table}

As in Table \ref{tab:human_eval}, INFO surpasses BART$_{base}$, BART$_{large}$, GPT-2$_{small}$ and GPT-2$_{medium}$ in all of the criteria. INFO achieves the highest rank in adequacy, fluency, and provenance and generates a more human-like response than other generative models. Also, the workers ranked our model the lowest when they were asked to rank the responses in the most hallucinated order. Thus, it can be found that INFO generates more engaging and fewer hallucination utterances with respect to the human. The distribution of the rank per each criterion is illustrated in Appendix \ref{sec:humaneval_dist}.

\section{Results and Analysis}
\subsection{Variants on Candidate Scoring Module}
To validate the poly-encoder as a candidate scoring module, we apply diverse candidate scoring modules, including the bi-encoder and cross-encoder. From the results in Table~\ref{tab:encoder_results}, we can find that the poly-encoder outperforms in the generation task. In the grounding task, $S_{RT}$ with cross-encoder scoring shows improved accuracy on grounding persona and knowledge. The result seems to be $S_{RT}$ with bi-encoder and cross-encoder are better than that with poly-encoder. However, the F1 score of INFO is higher than the two candidate scoring modules implying that low accuracy in persona is due to the tendency of active use on the persona in poly-encoder while the other two models opt to predict not to use persona sentence. The results suggest that the high accuracy of persona not always guarantees the engagingness in the dialogue.

\subsection{Comparison on other Retrievers}
We show that INFO is effective in retrieving knowledge compared to other sparse and dense retrievers. We retrieve the knowledge from our knowledge index built with Wikipedia paragraphs. We utilize TF-IDF~\citep{joachims1996probabilistic}, and deep passage retrieval (DPR)~\citep{karpukhin2020dense}. In the case of TF-IDF, we set the sum of query and knowledge tokens less than or equal to 512, which is the maximum sequence length of DPR and INFO. We use \texttt{bert-base-uncased} as the tokenizer. For DPR, we extract less than 40 knowledge using TF-IDF due to memory limitations. 
We first retrieve the five paragraphs related to the query that comprises knowledge snippet, dialogue history, predicted knowledge candidate, and selected persona sentences. In Table \ref{tab:results_retrievers},  we find that the retriever we used outperforms compared to the TF-IDF and DPR in all the metrics, including BERTscore. The results imply that INFO's retriever is suitable for extracting similar paragraphs rather than other retrievers. 

\begin{table}[h]
\centering
\scalebox{0.71}{
\begin{tabular}{lcccccc}
\toprule
\multicolumn{1}{l|}{Model} & chrF++ & BLEU & R-1 & R-2 & R-L & BERTScore \\ \midrule\midrule
\multicolumn{1}{l|}{TF-IDF} & 19.91 & 3.52 & 13.91 & 9.96 & 12.43 & 51.54  \\
\multicolumn{1}{l|}{DPR} & 20.57 & 3.86 & 12.44 & 6.55 & 10.20 & 47.48 \\ \cdashline{1-7}[0.8pt/1.2pt]
\multicolumn{1}{l|}{\textbf{INFO} }        & \textbf{26.36}                        & \textbf{7.40}               & \textbf{15.48}                        & \textbf{12.18}               & \textbf{14.32}        & \textbf{53.14}   \\ 
\bottomrule
\end{tabular}
}
\caption{Comparison with other retrievers\label{tab:results_retrievers}}
\end{table}

\subsection{Effect of Selectors on Generation}
We measure each selector module's effect on the generation task by changing the query which feds into the retriever on a validation set. The experimental results are shown in Table \ref{tab:gtp_gtk}, where $GT_K$, $GT_P$ represents ground truth knowledge and persona. Although the query that comprises the ground truth source shows the highest scores, INFO demonstrates comparable results on the generation task. From the result where the performance increase of INFO + $GT_P$ is larger than that of INFO + $GT_K$ about 2.8\%p, we can identify that our persona selector still has more space to achieve its maximum level.

\begin{table}[h]
\centering
\scalebox{0.65}{
\begin{tabular}{lcccccc}
\toprule
\multicolumn{1}{c|}{Query} & chrF++ & BLEU & R-1 & R-2 & R-L & BERTScore \\ \midrule
\midrule
\multicolumn{1}{l|}{\textbf{INFO} (RT)} & 53.29 & 31.46 & 58.26 & 42.35 & 53.06 & 92.29 \\
\cdashline{1-7}[0.8pt/1.2pt]
\multicolumn{1}{l|}{$+GT_K$} & 53.35 & 31.56 & 58.31 & 42.55 & 53.18 & 92.29                        \\
\multicolumn{1}{l|}{$+GT_P$} & 56.19 & 34.39 & 61.61 & 45.46 & 56.01 & 92.79                         \\
\multicolumn{1}{l|}{$+GT_K$+$GT_P$}        & \textbf{56.40}                        & \textbf{34.60}               & \textbf{61.88}                        & \textbf{45.64}               & \textbf{56.16}        & \textbf{92.84}                        \\ 

\bottomrule
\end{tabular}
}
\caption{Comparison between the generation performances based on the variants of query with ground truth knowledge and persona. Note that all the performance is evaluated with the official validation set. \label{tab:gtp_gtk}}
\end{table}

\begin{table}[t]
\centering
\scalebox{0.60}{
\begin{tabular}{ll}
\toprule
\multicolumn{2}{l}{\textbf{Given Landmark}} \\ \midrule
\multicolumn{2}{l}{\begin{tabular}[l]{p{11cm}}Finding Nemo Submarine Voyage\end{tabular}} \\ \midrule
\multicolumn{2}{l}{\textbf{Dialogue}} \\ \midrule
\multicolumn{2}{l}{\begin{tabular}[l]{p{11cm}}\textit{Human}:~What area of the park is this ride in? \\
\textit{Machine}:~This ride is located in the Tomorrowland area of Disneyland.\\
\textit{Human}:~Has this ride always been about Finding Nemo?\end{tabular}} \\ \midrule
\multicolumn{2}{l}{\textbf{Human's Persona}} \\ \midrule
\multicolumn{2}{l}{\begin{tabular}[l]{p{11cm}}
I've never been to California. \\
\textcolor{red}{My favorite cartoon is Finding Nemo.} \\
I would like to visit Disneyland.\\
My favorite color is yellow. \\
I enjoy swimming.\\ 
\end{tabular}} \\ 
\midrule
\multicolumn{2}{l}{\textbf{Ground Truth Knowledge (Grounding)}} \\ \midrule
\multicolumn{2}{l}{\begin{tabular}[l]{p{11cm}}Based on the characters and settings of the 2003 Disney·Pixar, \textcolor{blue}{Finding Nemo, it is a re-theming of the classic Submarine Voyage attraction that operated from 1959 to 1998.}\end{tabular}} \\
\midrule
\multicolumn{2}{l}{\textbf{Retrieved Knowledge} (Generation)} \\ \midrule
\multicolumn{2}{l}{\begin{tabular}[l]{p{11cm}}The original Submarine Voyage was built in 1959 as part of the then new Tomorrowland... \end{tabular}} \\
\cdashline{1-2}[0.8pt/1.2pt]
\multicolumn{2}{l}{\begin{tabular}[l]{p{11cm}}In 2008, Finding Nemo Submarine Voyage received an award for outstanding achievement from the Themed Entertainment Association. \end{tabular}} \\
\cdashline{1-2}[0.8pt/1.2pt]
\multicolumn{2}{l}{\begin{tabular}[l]{p{11cm}}Finding Nemo Submarine Voyage is ... which opened on June 11, 2007. ... it is a re-theming of the classic Submarine Voyage attraction that operated from 1959 to 1998. \end{tabular}} \\
\cdashline{1-2}[0.8pt/1.2pt]
\multicolumn{2}{l}{\begin{tabular}[l]{p{11cm}}On July 15, 2005, two days before the 50th Anniversary of Disneyland, the Finding Nemo Submarine Voyage was officially announced at the new Turtle Talk... \end{tabular}} \\
\cdashline{1-2}[0.8pt/1.2pt]
\multicolumn{2}{l}{\begin{tabular}[l]{p{11cm}}The attraction reuses the eight original 1959 Submarine Voyage through Liquid Space attraction... \end{tabular}} \\
\midrule
\multicolumn{2}{l}{\textbf{Predicted Answers}} \\ \midrule
\multicolumn{1}{l|}{~~BART$_{base}$} & \multicolumn{1}{l}{\begin{tabular}{p{7.75cm}}Yes, that's right. You're a fan of the ``Fantasy'' film, so I. \end{tabular}} \\
\cdashline{1-2}[0.8pt/1.2pt]
\multicolumn{1}{l|}{~~BART$_{large}$} & \multicolumn{1}{l}{\begin{tabular}{p{7.75cm}}Yes, the ride is based on the characters and settings of the 2003 Disney·Pixar film \end{tabular}} \\
\cdashline{1-2}[0.8pt/1.2pt]
\multicolumn{1}{l|}{~~GPT-2$_{small}$} & \multicolumn{1}{l}{\begin{tabular}{p{7.75cm}}No, it was originally a way to show that you love Finding Nemo.\end{tabular}} \\
\cdashline{1-2}[0.8pt/1.2pt]
\multicolumn{1}{l|}{~~GPT-2$_{medium}$} & \multicolumn{1}{l}{\begin{tabular}{p{7.75cm}}Yes, it has operated from 1959 to 1998.\end{tabular}}  \\
\midrule
\multicolumn{1}{l|}{~~\textbf{INFO} (Ours)} & \multicolumn{1}{l}{\begin{tabular}{p{7.75cm}}No, \textcolor{blue}{this attraction is actually a re-theme of the classic submarine voyage attraction that operated from 1959 to 1998. The attraction is based on the characters and settings of the 2003 Disney Pixar film Finding Nemo,} \textcolor{red}{which is your favorite cartoon.}\end{tabular}} \\ 
\midrule
\multicolumn{2}{l}{\textbf{Ground Truth Response}} \\
\midrule
\multicolumn{2}{l}{\begin{tabular}[l]{p{11cm}}No, your favorite cartoon is a new addition to this ride. The current Finding Nemo ride is a re-theming of the classic ``Submarine Voyage'' attraction that operated here from 1959 to 1998.\end{tabular}}  \\ 
\bottomrule 
\end{tabular}}
\caption{Qualitative result. All the predicted results in grounding task are from our model, INFO and it predicts the correct answers in both tasks. We add other baselines' responses for comparative analysis.\label{tab:qualitative}}
\end{table}

\subsection{Qualitative Analysis}
In Table \ref{tab:qualitative}, an example from the predicted results is illustrated. In the case of BART$_{large}$, and GPT-2$_{medium}$, the responses only reflect the ground truth knowledge resulting in less engaged answers without any persona-related phrases. Although BART$_{base}$ seems to employ a persona sentence in the form of the phrase ``You're fan of the Fantasy film'', its used sentence does not appear in human's personal profiles. This result also indicates that the utterance is hard to identify its provenance on the knowledge source. Moreover, GPT-2$_{small}$ generates the utterance that contradicts the ground truth knowledge. From the result, we can find that the generated responses from the baselines show hallucinations on both persona and knowledge. Unlike other baselines, our model blends ground truth knowledge and persona sentence into the response with less hallucination and engagingness. In addition, the retrieved knowledge source that our model refers to provides interpretability and provenance of the responses to the users. More examples are also depicted in Appendix \ref{sec:appen_qual}.

\section{Conclusions}
In this paper, we presented a conversational agent that generates responses grounding the user's persona and external knowledge. We 
utilized poly-encoder-based candidate scoring for each grounding task. We additionally implement persona level indicator to consider multiple persona selections for delicate persona grounding. With predicted sources, we construct a knowledge-persona enhanced query to retrieve latent paragraphs, and they are used to generate informative and engaging responses by marginalizing loss for each token. We show that our method achieves the state-of-the-art (SoTA) score in both grounding and generation tasks in the persona-knowledge conversation dataset. We also demonstrate that the responses from INFO show less hallucination and more engagingness through human evaluation and qualitative analysis. We also compare the grounding modules and retrievers to show INFO's effectiveness. 

\section{Limitations}
The proposed model INFO has limitations. Given the INFO's settings, the model cannot deal with real-world application, which means the absence of ground truth knowledge or persona candidates in the grounding task. We also conducted the human evaluation to evaluate the capability of the proposed model's mitigating hallucination in dialogue generation. However, the number of cases is relatively small for evaluating the capability of mitigating hallucination. Finally, INFO demands high GPU computation resources, since it marginalizes loss at the token level. 

We plan to improve the INFO for future work. We will train and evaluate the INFO in open-domain settings as well as real-world settings for the applicable conversational agents. Moreover, we will conduct human evaluations with more cases. Especially, we will enhance the way of quantitative measurement for the model's hallucinated answers. Last but not least, we will improve the generator of INFO with more computationally efficient components.

\section{Acknowledgement}
This work was supported by Institute of Information \& communications Technology Planning \& Evaluation(IITP) grant funded by the Korea government(MSIT) (No. 2020-0-00368, A Neural-Symbolic Model for Knowledge Acquisition and Inference Techniques),  This research was supported by the MSIT(Ministry of Science and ICT), Korea, under the ITRC(Information Technology Research Center) support program(IITP-2022-2018-0-01405) supervised by the IITP(Institute for Information \& Communications Technology Planning \& Evaluation), This work was supported by Institute for Information \& communications Technology Planning \& Evaluation(IITP) grant funded by the Korea government(MSIT) (No. 2022-0-00369, (Part 4) Development of AI Technology to support Expert Decision-making that can Explain the Reasons/Grounds for Judgment Results based on Expert Knowledge)

\bibliography{emnlp2022_arxiv}
\bibliographystyle{acl_natbib}
\clearpage
\appendix

\onecolumn
\centering
\section{Automatic Evaluation on Official Test Set} \label{sec:appen_main}

\begin{table*}[h]
\centering
\scalebox{0.80}{
\begin{tabular}{lcccccccc}
\toprule
\multicolumn{1}{c|}{\multirow{2}{*}{Models}} & \multicolumn{5}{c|}{Generation}                       & \multicolumn{2}{c}{Grounding (Acc.)} \\ \cline{2-8} 
\multicolumn{1}{c|}{}  
& \multicolumn{1}{c}{chrF++}  & \multicolumn{1}{c}{BLEU} & \multicolumn{1}{c}{R-1} & \multicolumn{1}{c}{R-2} & \multicolumn{1}{c|}{R-L}  & \multicolumn{1}{c}{Persona}          & \multicolumn{1}{c}{Knowledge}         \\ \midrule
\midrule

\multicolumn{1}{l|}{GPT2$_{small}$} & 28.83 & 11.60 & 36.28 & 19.56 & 32.42 & \multicolumn{1}{|c}{67.83} & 70.95 \\
\multicolumn{1}{l|}{GPT2$_{medium}$} & 30.34 & 12.58 & 38.35  & 21.16 & 34.34  & \multicolumn{1}{|c}{67.64} & 72.46 \\ \midrule
\multicolumn{1}{l|}{BART$_{base}$} & 29.80 & 12.15 & 36.26 & 19.73 & 32.06  &  \multicolumn{1}{|c}{67.66} & 72.02 \\
\multicolumn{1}{l|}{BART$_{large}$} & 30.63 & 11.86 & 36.36 & 19.42 & 31.73  & \multicolumn{1}{|c}{67.62} & 70.53 \\ \midrule

\multicolumn{1}{l|}{\textbf{INFO} (RS)} & 52.81 & 29.41 & 56.37 & 40.41 & 51.16 & \multicolumn{1}{|c}{\textbf{82.74}} & \multicolumn{1}{c}{98.88} \\
\multicolumn{1}{l|}{\textbf{INFO} (RT)} & \textbf{54.61} & \textbf{32.33} & \textbf{58.27} & \textbf{42.39} & \textbf{53.09}  & \multicolumn{1}{|c}{80.83} & \multicolumn{1}{c}{\textbf{99.10}} \\

\bottomrule
\end{tabular}
}
\caption{Main results on the official test set. RT indicates the model with RAG-Token model as generator. The models are evaluated by generation metrics, including chrF++, BLEU, ROUGE-1 (R-1), ROUGE-2 (R-2) and ROUGE-L (R-L). The accuracy for persona grounding task and knowledge grounding task are also noted. Since BERTscore is not the official generation metric, we cannot evaluate the result on the metric as the ground truth of the test is not yet disclosed. \label{tab:appen_main_results}}
\end{table*}

\section{Human Evaluation Distribution on Each Criteria} \label{sec:humaneval_dist}

\begin{figure}[ht]
\begin{subfigure}[b]{.49\linewidth}
\centering
\includegraphics[width=\linewidth]{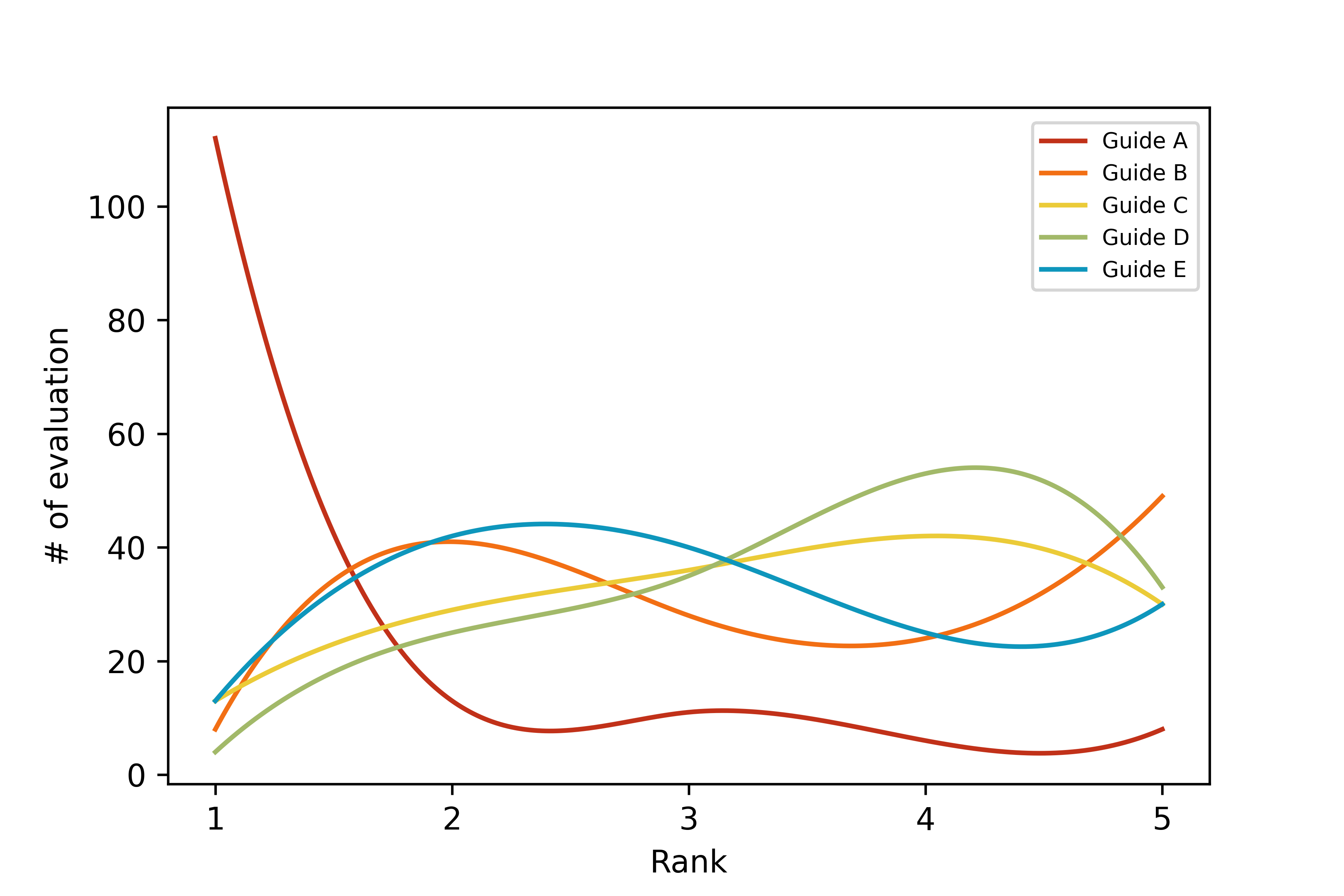}
\caption{Adequacy}
\label{fig:human_a}
\end{subfigure}\hfill
\begin{subfigure}[b]{.49\linewidth}
\centering
\includegraphics[width=\linewidth]{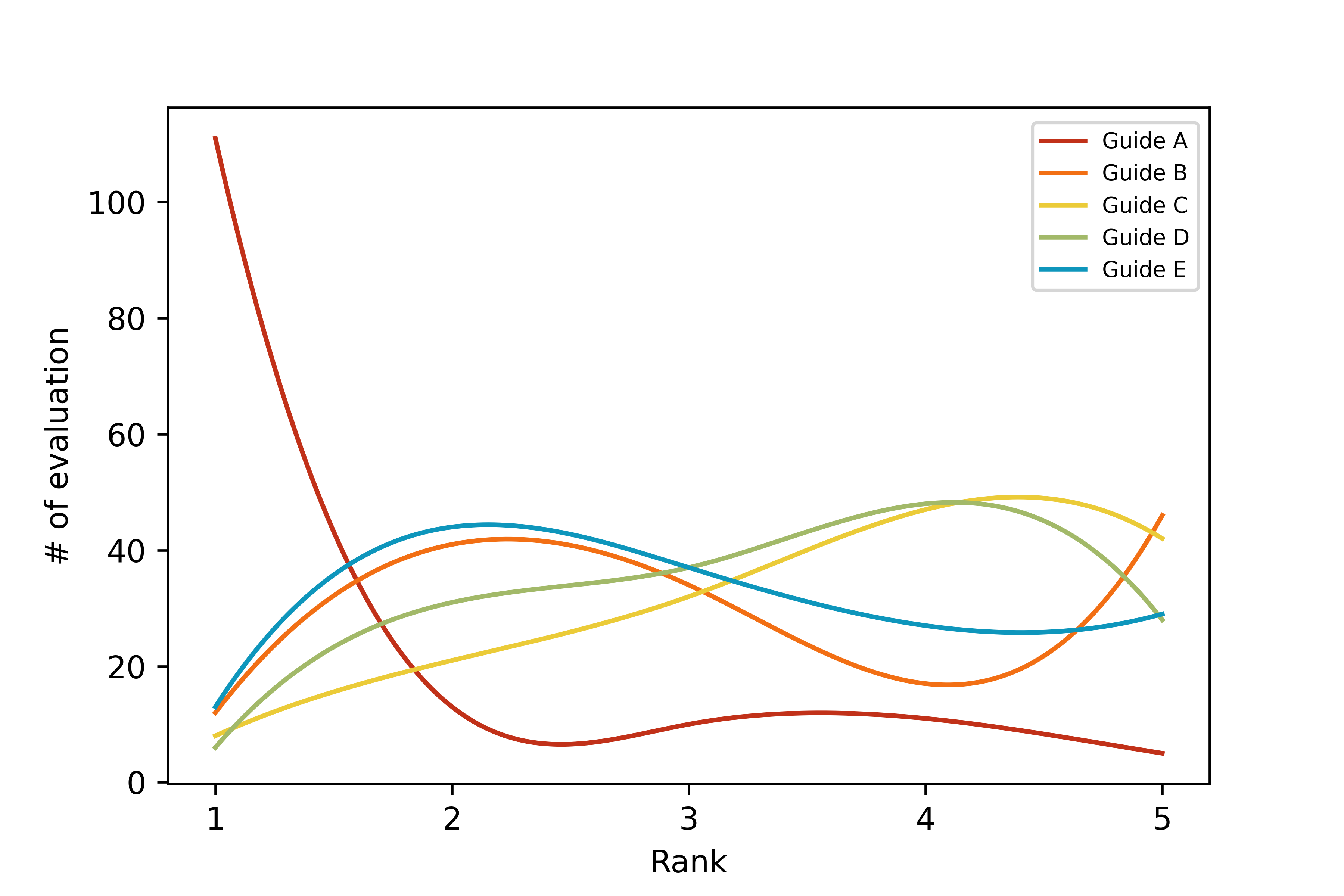}
\caption{Fluency}
\label{fig:human_f}
\end{subfigure}%
\caption{The distribution of the rank on the adequacy and fluency criteria. Guide A to E indicates INFO, BART$_{base}$, BART$_{large}$, GPT-2$_{small}$, and GPT-2$_{medium}$, in the order.}
\label{fig:human_af}
\end{figure}

\begin{figure}[ht]
\begin{subfigure}[b]{.49\linewidth}
\centering
\includegraphics[width=\linewidth]{figures/graph_adequate.png}
\caption{Provenance}
\label{fig:human_p}
\end{subfigure}\hfill
\begin{subfigure}[b]{.49\linewidth}
\centering
\includegraphics[width=\linewidth]{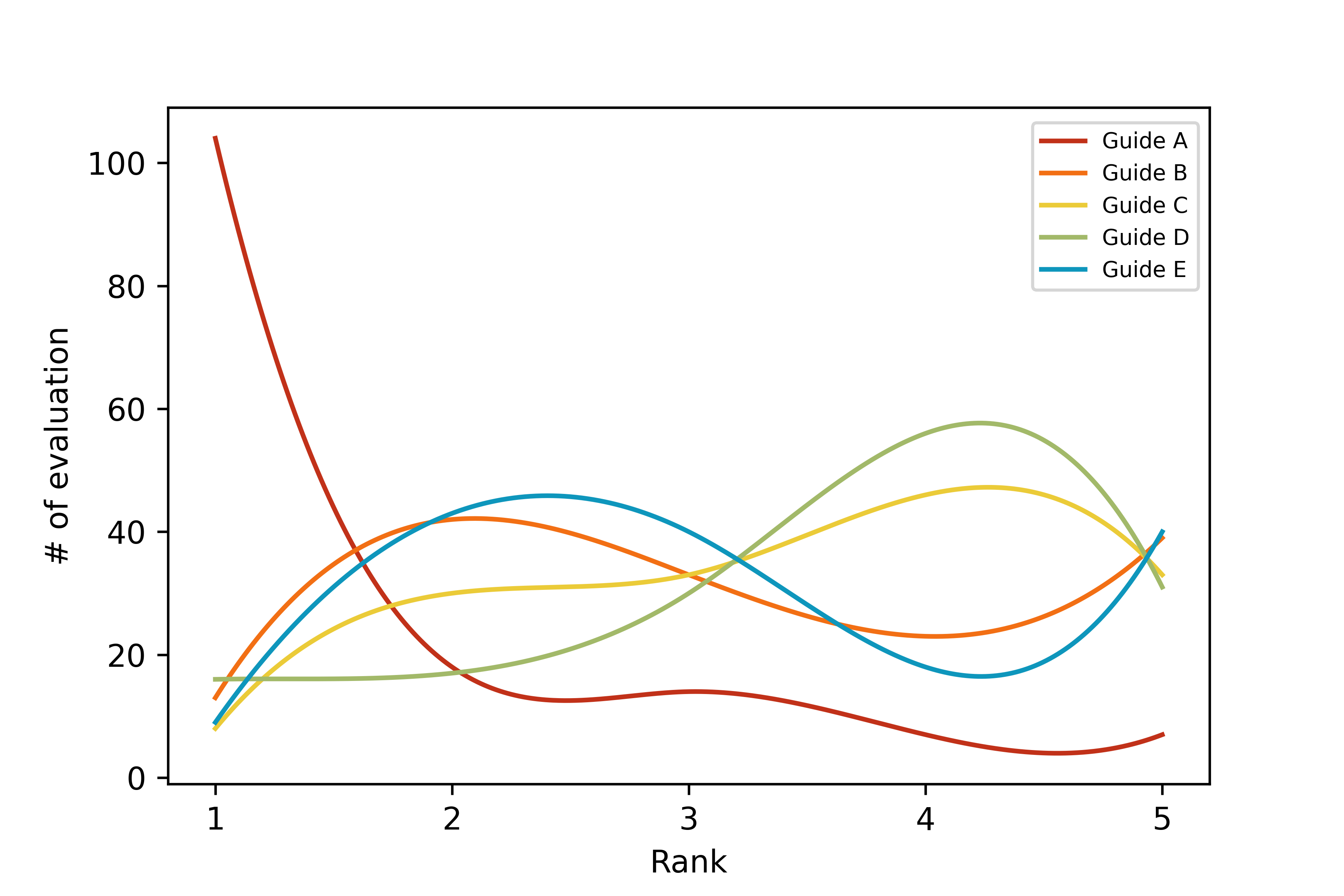}
\caption{Engagingness}
\label{fig:human_e}
\end{subfigure}%
\caption{The distribution of the rank on the provenance and engagingness criteria. Guide A to E indicates INFO, BART$_{base}$, BART$_{large}$, GPT-2$_{small}$, and GPT-2$_{medium}$, in the order.}
\label{fig:human_pe}
\end{figure}

\begin{figure*}[ht]
	\centering
	\includegraphics[width=0.49\textwidth]{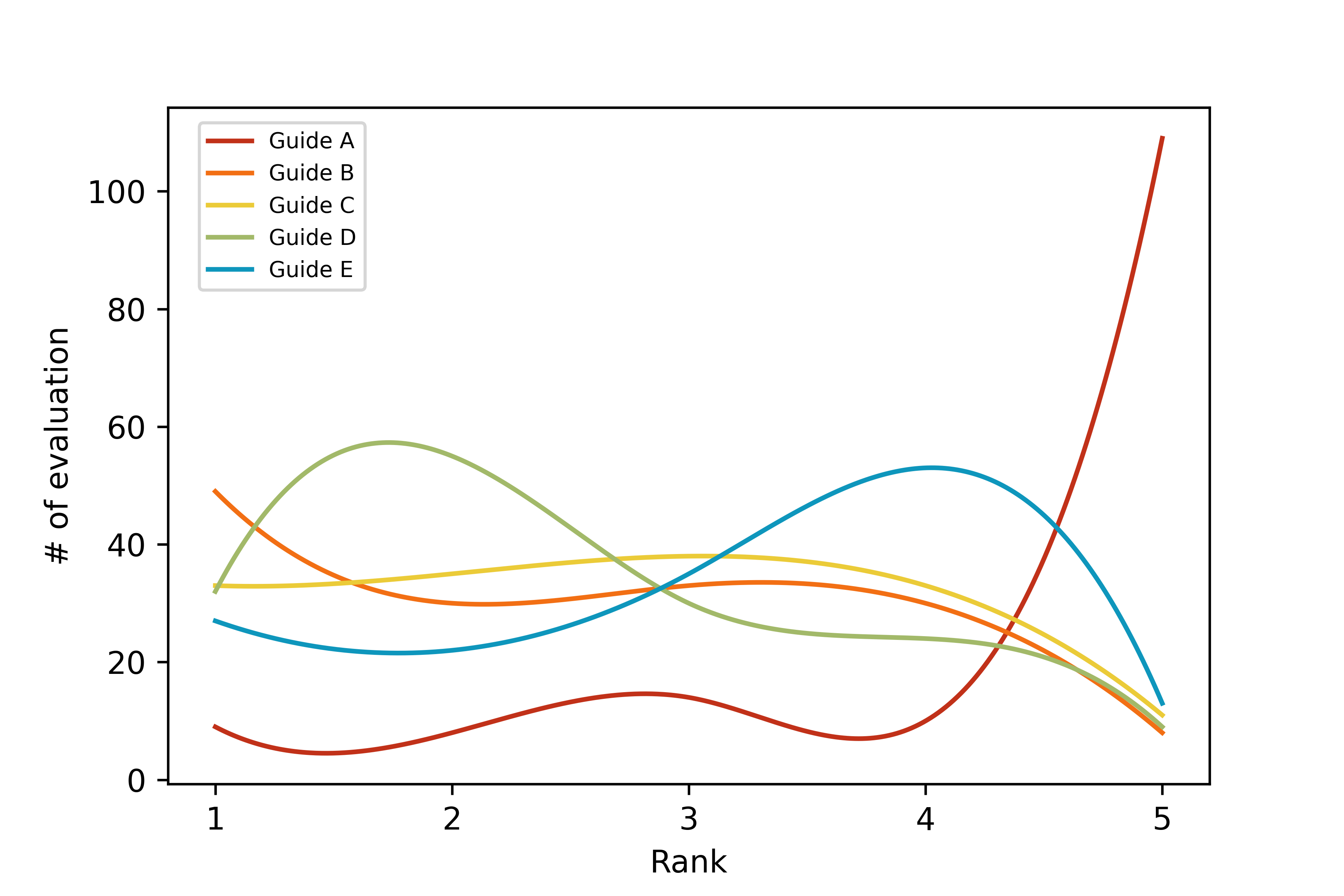}
\caption{The distribution of the rank on the less hallucination criterion. Note that the highest rank (1) means the most hallucinated. Guide A to E indicates INFO, BART$_{base}$, BART$_{large}$, GPT-2$_{small}$, and GPT-2$_{medium}$, in the order. } \label{fig:human_h}
\end{figure*}

\clearpage

\section{Qualitative Results} 
\label{sec:appen_qual}

\begin{table}[ht]
\centering
\scalebox{0.64}{
\begin{tabular}{ll}
\toprule
\multicolumn{2}{l}{\textbf{Given Landmark}} \\ \midrule
\multicolumn{2}{l}{\begin{tabular}[l]{p{11cm}}Nocton Hall\end{tabular}} \\ \midrule
\multicolumn{2}{l}{\textbf{Dialogue}} \\ \midrule
\multicolumn{2}{l}{\begin{tabular}[l]{p{11cm}}\textit{Human}:~I know this place, but I don't remember the name of this place. \end{tabular}} \\ \midrule
\multicolumn{2}{l}{\textbf{Human's Persona}} \\ \midrule
\multicolumn{2}{l}{\begin{tabular}[l]{p{11cm}}
I have a hall in my house.\\
I have worked with military hospital.\\
\textcolor{red}{I would like to go to England.}\\
I have heard about Frederick John Robinson.\\
I have respect to American Officers. 
\end{tabular}} \\ 
\midrule
\multicolumn{2}{l}{\textbf{Ground Truth Knowledge (Grounding)}} \\ \midrule
\multicolumn{2}{l}{\begin{tabular}[l]{p{11cm}}\textcolor{blue}{Today the site is maintained by English Heritage.}\end{tabular}} \\
\midrule
\multicolumn{2}{l}{\textbf{Retrieved Knowledge (Generation)}} \\ \midrule
\multicolumn{2}{l}{\begin{tabular}[l]{p{11cm}}``Hogesdon'' is first recorded in the Domesday Book, meaning an Anglo-Saxon farm (or ``fortified enclosure'') belonging to Hoch, or Hocq.... \end{tabular}} \\
\cdashline{1-2}[0.8pt/1.2pt]
\multicolumn{2}{l}{\begin{tabular}[l]{p{11cm}}By 1677, the church was in such poor structural condition that the Dean of Chichester convened a consistory court with three parishioners and asked them why repairs had not been carried out...\end{tabular}} \\
\cdashline{1-2}[0.8pt/1.2pt]
\multicolumn{2}{l}{\begin{tabular}[l]{p{11cm}}he south transept had historically served as the mausoleum for the Cavendish family, who resided at nearby Latimer... \end{tabular}} \\
\cdashline{1-2}[0.8pt/1.2pt]
\multicolumn{2}{l}{\begin{tabular}[l]{p{11cm}}Following the diminished influence of the ancient Priory, predecessor of the two parishes of St Bartholomew, disputes began to arise over rights to tithes and taxes payable by lay residents who claimed allegiance with the nearby and anciently associated parish of St Botolph Aldersgate - an unintended consequence and legacy of King Kenry VIII's religious reforms.\end{tabular}} \\
\cdashline{1-2}[0.8pt/1.2pt]
\multicolumn{2}{l}{\begin{tabular}[l]{p{11cm}}a wall-mounted stone pulpit, a stone altar, a series of tall, pointed windows high in the walls, an unadorned stone font and a short wooden steeple-little more than an extended belfry-extending from the nave roof...a wall-mounted stone pulpit, a stone altar, a series of tall, pointed windows high in the walls, an unadorned stone font and a short wooden steeple-little more than an extended belfry-extending from the nave roof... \end{tabular}} \\
\midrule
\multicolumn{2}{l}{\textbf{Predicted Answers}}  \\ \midrule
\multicolumn{1}{l|}{~~BART$_{base}$} & \multicolumn{1}{l}{\begin{tabular}{p{7.75cm}}This is the Nocton Hall, a historic Grade II listed building in the village of Nocton, \end{tabular}} \\
\cdashline{1-2}[0.8pt/1.2pt]
\multicolumn{1}{l|}{~~BART$_{large}$} & \multicolumn{1}{l}{\begin{tabular}{p{7.75cm}}Nocton Hall is a historic Grade II listed building in the village of Nocton, in Lincoln \end{tabular}} \\
\cdashline{1-2}[0.8pt/1.2pt]
\multicolumn{1}{l|}{~~GPT-2$_{small}$} & \multicolumn{1}{l}{\begin{tabular}{p{7.75cm}}This is Nocton Hall, a historic Grade II listed building in the village of Nocton, in \end{tabular}} \\
\cdashline{1-2}[0.8pt/1.2pt]
\multicolumn{1}{l|}{~~GPT-2$_{medium}$} & \multicolumn{1}{l}{\begin{tabular}{p{7.75cm}}This is the Nocton Hall, a historic Grade II listed building in the village of Nocton,\end{tabular}} \\
\cdashline{1-2}[0.8pt/1.2pt]
\multicolumn{1}{l|}{~~\textbf{INFO} (Ours)} & \multicolumn{1}{l}{\begin{tabular}{p{7.75cm}}This will be a nice vision for you. \textcolor{blue}{Nocton Hall is a historic Grade II listed building in Eng}\textcolor{red}{land that you would like to visit}.\end{tabular}} \\ 
\midrule
\multicolumn{2}{l}{\textbf{Ground Truth Response}}                             \\ \midrule
\multicolumn{2}{l}{\begin{tabular}[l]{p{11cm}}You can see Nocton Hall in the village of Nocton, in Lincolnshire of England, the country you want to visit.\end{tabular}} \\ 
\bottomrule 
\end{tabular}}
\centering
\scalebox{0.64}{
\begin{tabular}{ll}
\toprule
\multicolumn{2}{l}{\textbf{Given Landmark}} \\ \midrule
\multicolumn{2}{l}{\begin{tabular}[l]{p{11cm}}Maiden Castle, Dorset\end{tabular}} \\ \midrule
\multicolumn{2}{l}{\textbf{Dialogue}} \\ \midrule
\multicolumn{2}{l}{\begin{tabular}[l]{p{11cm}}\textit{Human}:~Wow, this is amazing! What is this?\\
\textit{Machine}:~It is Maiden Castle in Dorset. I thought you would like it since\\
~~~~~~~~~~~~~~~you are interested in historic forts.\\
\textit{Human}:~Who owns the site today?\end{tabular}}
\\ \midrule
\multicolumn{2}{l}{\textbf{Human's Persona}} \\ \midrule
\multicolumn{2}{l}{\begin{tabular}[l]{p{11cm}}
I like Britain.\\
I have been to Dorset.\\
I am interested in historic forts.\\
\textcolor{red}{I hope to work for English Heritage.}\\
I would like to visit an old fort. 
\end{tabular}} \\ 
\midrule
\multicolumn{2}{l}{\textbf{Ground Truth Knowledge (Grounding)}} \\ \midrule
\multicolumn{2}{l}{\begin{tabular}[l]{p{11cm}}\textcolor{blue}{Today the site} is protected as a Scheduled Ancient Monument and \textcolor{blue}{is maintained by English Heritage}.\end{tabular}} \\\midrule
\multicolumn{2}{l}{\textbf{Retrieved Knowledge} (Generation)} \\ \midrule
\multicolumn{2}{l}{\begin{tabular}[l]{p{11cm}}Portland Castle is an artillery fort constructed by Henry VIII on the Isle of Portland, Dorset, between 1539 and 1541... \end{tabular}} \\
\cdashline{1-2}[0.8pt/1.2pt]
\multicolumn{2}{l}{\begin{tabular}[l]{p{11cm}}this version of events, or even that the hill fort was attacked by the Romans... \end{tabular}} \\
\cdashline{1-2}[0.8pt/1.2pt]
\multicolumn{2}{l}{\begin{tabular}[l]{p{11cm}}Between 1985 and 1986 further excavations under Niall Sharples were prompted by the hill fort's deteriorating condition, partly caused by the large number of visitors to the site... \end{tabular}} \\
\cdashline{1-2}[0.8pt/1.2pt]
\multicolumn{2}{l}{\begin{tabular}[l]{p{11cm}}a Tudor rose and the initials E.R. (Elizabeth Regina), has been preserved and can be seen in the inner bailey of the castle mounted on a replica carriage... \end{tabular}} \\
\cdashline{1-2}[0.8pt/1.2pt]
\multicolumn{2}{l}{\begin{tabular}[l]{p{11cm}}Constructed on a territorial boundary in about 600 BC, the first hill fort at Maiden Castle was a 6.4-hectare (16-acre) area surrounded by a single ditch... \end{tabular}} \\

\multicolumn{2}{l}{\textbf{Predicted Answers}}  \\ \midrule
\multicolumn{1}{l|}{~~BART$_{base}$} & \multicolumn{1}{l}{\begin{tabular}{p{7.75cm}}The site is maintained by English Heritage, the country you are from.  \end{tabular}} \\
\cdashline{1-2}[0.8pt/1.2pt]
\multicolumn{1}{l|}{~~BART$_{large}$} & \multicolumn{1}{l}{\begin{tabular}{p{7.75cm}}Today the site is owned by English Heritage..... \end{tabular}} \\
\cdashline{1-2}[0.8pt/1.2pt]
\multicolumn{1}{l|}{~~GPT-2$_{small}$} & \multicolumn{1}{l}{\begin{tabular}{p{7.75cm}}Today the site is protected as a Scheduled Ancient Monument and is maintained by English Heritage. \end{tabular}} \\
\cdashline{1-2}[0.8pt/1.2pt]
\multicolumn{1}{l|}{~~GPT-2$_{medium}$} & \multicolumn{1}{l}{\begin{tabular}{p{7.75cm}}Today the site is maintained by English Heritage. \end{tabular}} \\
\cdashline{1-2}[0.8pt/1.2pt]
\multicolumn{1}{l|}{~~\textbf{INFO} (Ours)} & \multicolumn{1}{l}{\begin{tabular}{p{7.75cm}}\textcolor{blue}{Today the site is owned by English Heritage}. You may wish to research this further since \textcolor{red}{you hope to work for English Heritage}.\end{tabular}} \\ 
\midrule
\multicolumn{2}{l}{\textbf{Ground Truth Response}} \\
\midrule
\multicolumn{2}{l}{\begin{tabular}[l]{p{11cm}}It is owned by English Heritage; a company you hope to work for.\end{tabular}}  \\ 
\bottomrule 
\end{tabular}}

\caption{Qualitative results. All the predicted results in grounding task are from our model, INFO and it predicts the correct answers in both tasks. We add other baselines' responses for comparative analysis.\label{tab:qualitative3}}
\end{table}
\end{document}